\documentclass[letterpaper, 10 pt, journal, twoside]{IEEEtran}
\IEEEoverridecommandlockouts 
\usepackage{cite}
\usepackage{graphics} 
\usepackage{epsfig} 
\usepackage{mathptmx} 
\usepackage{times} 
\usepackage{amssymb}  
\usepackage[cmex10]{amsmath}
\usepackage{amsthm}
\usepackage{amsfonts}
\usepackage{latexsym}
\usepackage{mathrsfs}
\usepackage{xcolor}
\usepackage[normalem]{ulem}
\usepackage{stfloats}
\usepackage{mathtools}
\usepackage{wrapfig}
\usepackage{xspace}
\usepackage{pbox}
\usepackage{graphicx}
\usepackage{placeins}
\usepackage{flushend}

\ifCLASSOPTIONcompsoc
    \usepackage[caption=false, font=normalsize, labelfont=sf, textfont=sf]{subfig}
\else
\usepackage[caption=false, font=footnotesize]{subfig}
\fi

\theoremstyle{remark}

\theoremstyle{definition}
\newtheorem{defn}{Definition}

\theoremstyle{plain}

\newtheorem{thm}{Theorem}
\newtheorem{prop}{Proposition}

\newcommand{\newsec}[1]{\noindent \textbf{#1.} }

\definecolor{wdc}{RGB}{34,139,34}

\definecolor{cal}{RGB}{255,110,30}

\begin{document}


\title{Towards Variable Assistance for Lower Body Exoskeletons}

\author{Thomas Gurriet$^{1}$, Maegan Tucker$^{1}$, Alexis Duburcq$^{2}$,  Guilhem Boeris$^{2}$ and Aaron D. Ames$^{1}$%
\thanks{Manuscript received: 09, 11, 2019; Accepted 11, 06, 2019.}
\thanks{This paper was recommended for publication by Pietro Valdastri upon evaluation of the Associate Editor and Reviewers' comments.}
\thanks{This work was supported by NSF NRI award 1724464, the Caltech Big Ideas and ZEITLIN Funds, and Wandercraft. This work was conducted under IRB No. 16-0693.}%
\thanks{$^{1}$Thomas Gurriet, Maegan Tucker and Aaron Ames are with the Department
of Mechanical and Civil Engineering, California Institute of Technology, Pasadena, California {\tt\footnotesize \{tgurriet,mtucker,ames\}@caltech.edu}.}%
\thanks{$^{2}$Alexis Duburcq and Guilhem Boeris are with Wandercraft, Paris, France {\tt\footnotesize \{guilhem.boeris,alexis.duburcq\}@wandercraft.eu}.}
\thanks{Digital Object Identifier (DOI): see top of this page.}
}

\markboth{IEEE Robotics and Automation Letters. Preprint Version. Accepted 11, 2019}
{Gurriet \MakeLowercase{\textit{et al.}}: Variable Assistance for Lower Body Exoskeletons}

\maketitle

\begin{abstract}

This paper presents and experimentally demonstrates a novel framework for variable assistance on lower body exoskeletons, based upon safety-critical control methods. Existing work has shown that providing some freedom of movement around a nominal gait, instead of rigidly following it, accelerates the spinal learning process of people with a walking impediment when using a lower body exoskeleton. With this as motivation, we present a method to accurately control how much a subject is allowed to deviate from a given gait while ensuring robustness to patient perturbation. This method leverages control barrier functions to force certain joints to remain inside predefined trajectory tubes in a minimally invasive way. The effectiveness of the method is demonstrated experimentally with able-bodied subjects and the Atalante lower body exoskeleton.

\end{abstract}


\IEEEpeerreviewmaketitle

\begin{figure*}[b!]
    \centering
    \includegraphics[width=\textwidth]{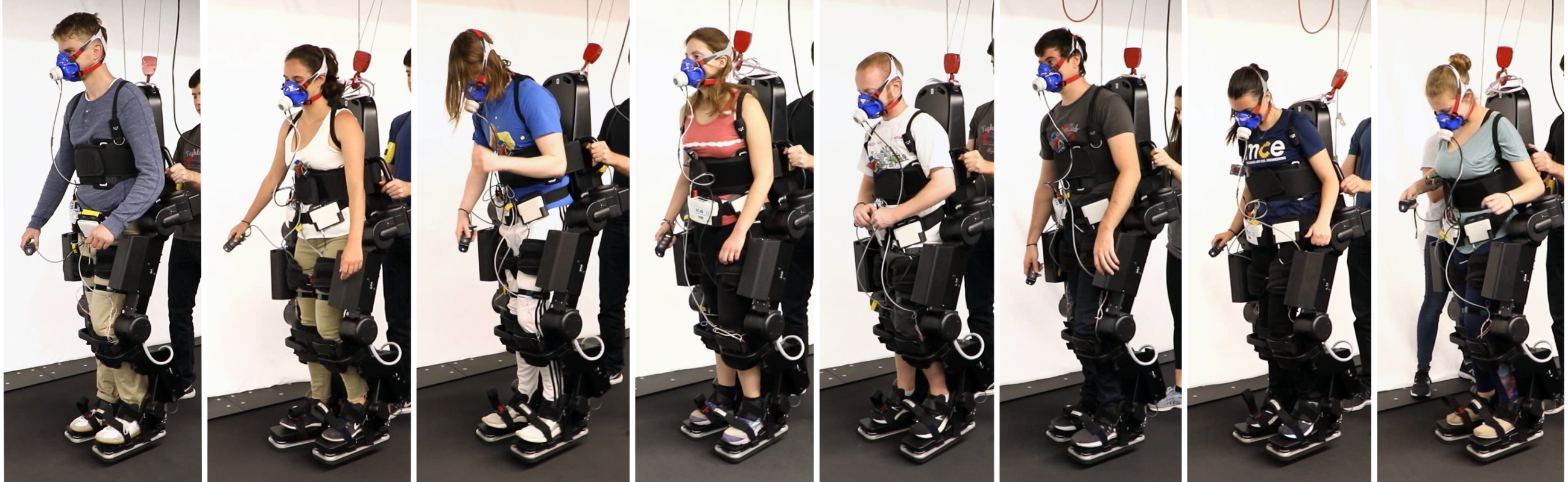}
    \caption{Photos of the eight able-bodied subjects who participated in the experimental evaluation.}
    \label{fig:introPicture}
\end{figure*}

\section{Introduction}
\label{sec:intro}

\IEEEPARstart{A}{ctive} lower-limb exoskeleton technology has the potential to benefit approximately 6.4 million people in the United States who are limited by the effects of stroke, polio, multiple sclerosis, spinal cord injury, and cerebral palsy \cite{HerrActiveOrthoses}. The term ``exoskeleton'' is traditionally associated with devices that assist people with physical disabilities \cite{chen2015design, esquenazi2012rewalk, neuhaus2011design, swift2011control}. Additionally, exoskeletons can also be designed to improve strength and endurance of able-bodied persons \cite{krut2010moonwalker, zoss2006biomechanical}.

The main focus of this paper is exoskeleton technology aimed at restoring locomotion for people with a leg pathology. While mechanical design is an important consideration for the development of exoskeleton devices, this paper focuses on the control methodology. A general review of control strategies for lower-limb assistive devices is given in \cite{jimenez2012review, tucker2015control, anam2012active}. Most current approaches to control powered leg devices are driven by finite-state machines with each phase defined using heuristic parameters. This approach typically requires the use of additional stability aids such as arm-crutches. Recently, dynamically stable crutch-less exoskeleton walking has been demonstrated for patients with paraplegia by leveraging the full nonlinear dynamics of the system and generating dynamically stable gaits \cite{gurriet2018towards}. The exoskeleton is then driven to follow these fixed trajectories.  

While this full assistance approach enables crutch-less exoskeleton walking, it is no longer optimal when exoskeleton technology is extended to patients who are recovering muscle functionality. For patients who are trying to strengthen recovering muscles, partial assistance would be more appropriate than full assistance. A previous study showed that permitting partial assistance and variability during step training enhanced stepping recovery after a complete spinal cord transection in adult mice \cite{cai2006implications}. The study also hypothesized that a fixed trajectory training strategy would drive the spinal circuitry toward a state of learned helplessness. These "assist-as-needed" algorithms, which have also been explored in other publications \cite{srivastava2014assist, cai2006assist, zanotto2014adaptive}, utilize velocity field control to provide gentle guidance at a constant rate towards the desired walking trajectory. 

The algorithm presented in this paper proposes a complementary approach that leverages tools from controlled set invariance \cite{aubin2009viability,blanchini2008set} -- in particular, control barrier functions \cite{ames2016control, ames2019control} -- to enable assist-as-needed strategies while guaranteeing coherence of the walking pattern. The method allows users to control their own motions when they are performing well (i.e. staying in a tube around a nominal trajectory) but intervene when they are not, so as to maintain a functional walking pattern. 
This approach, therefore, takes motivation from the growing area of safety-critical control \cite{ames2019control, agrawal2017discrete, wang2017safety}, and extends its application to exoskeletons with experimental demonstration with multiple subjects. 

In summary, this paper proposes a variable assistance framework targeted for patients who are in the process of recovering muscle functionality. Sec. \ref{sec:theory} discusses the mathematical theory behind the variable assistance framework. Sec. \ref{sec:results} presents and discusses the experimental results. Lastly, Sec. \ref{sec:conclusion} discusses the conclusions.

\section{Variable Assistance Framework}
\label{sec:problemFormulation}

\subsection{The Atalante Exoskeleton Platform}
\label{subsec:hardware}

\begin{figure}[t]
    \centering
    \includegraphics[width=0.9\columnwidth]{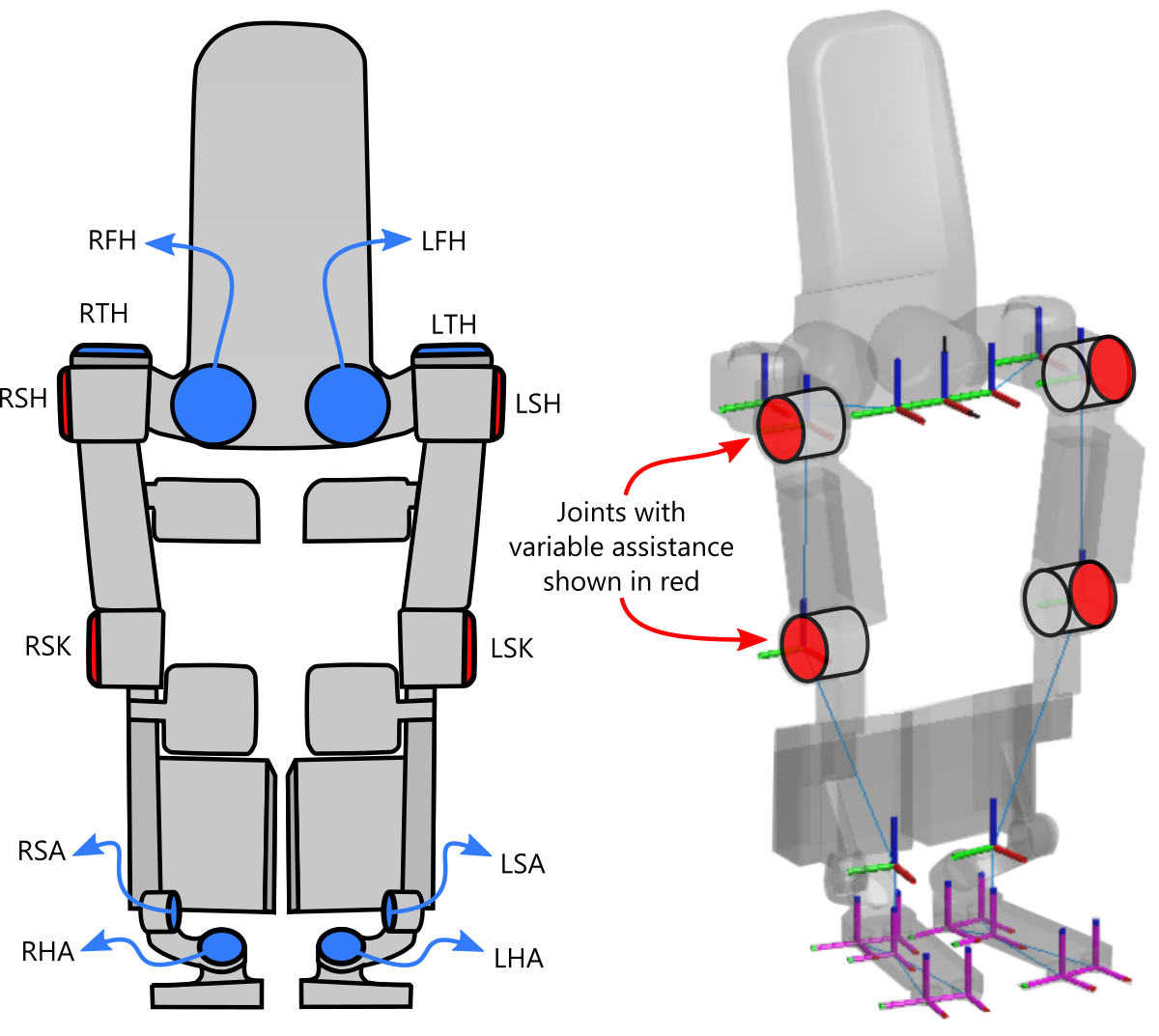}
    \caption{Image of the hardware. Silhouette with joints colored. In red are the joints that will be used for variable assistance.}
    \label{fig:hardwarePicture}
\end{figure}

The exoskeleton used for this work, named Atalante, was developed by the French startup company Wandercraft and has already demonstrated its ability to perform crutch-less dynamic walking with patients with paraplegia \cite{gurriet2018towards}. As shown in Fig. \ref{fig:hardwarePicture}, Atalante has a total of 12 actuated joints. Each leg has three actuated joints at the hip which control the spherical motion of the hip, one actuated joint at the knee, and two actuated joints at the ankle. The terms and abbreviations for the joints are as follows: frontal hip (FH), transverse hip (TH), sagittal hip (SH), sagittal knee (SK), sagittal ankle (SA), and henke ankle (HA). As for the sensing capabilities of Atalante, the position and velocity of each joint is measured using a digital encoder mounted on the motor. Additional attitude estimations are obtained using four Inertial Measurement Units (IMUs) that are located on the torso, the pelvis, the left tibia, and the right tibia. Finally, ground force information is obtained using eight 3-axis force sensors, four located on the bottom of each foot. The controller is run on a central processing unit running a real-time operating system.

\begin{figure}[t]
    \centering
    \includegraphics[width=\columnwidth]{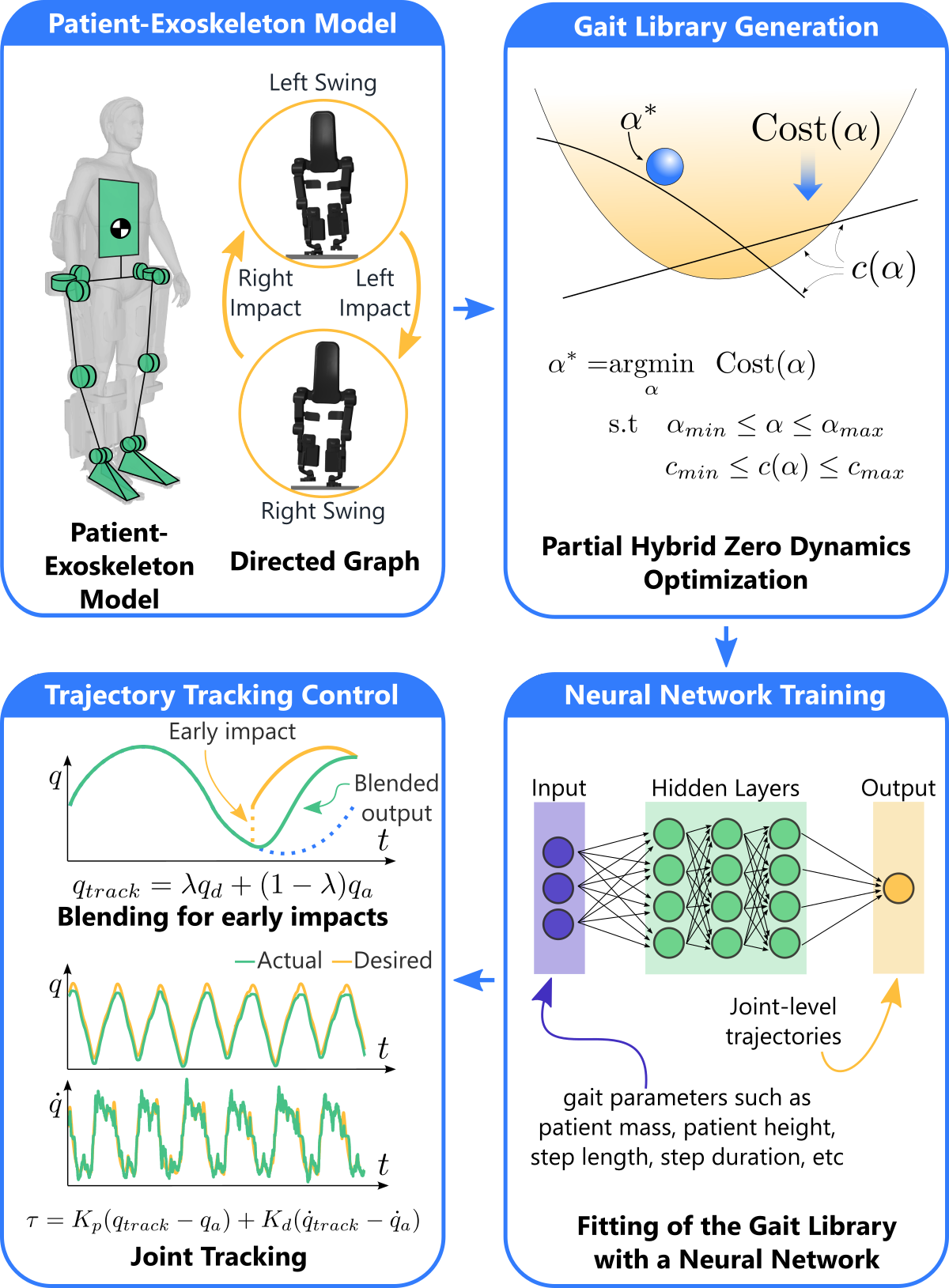}
    \caption{Structure of the gait library generation and joint tracking.}
    \label{fig:gaitLibArchitecture}
\end{figure}

\subsection{Baseline Walking Approach}
\label{subsec:baselineWalking}
The baseline walking approach used in this work, on which the variable assistance framework is added, consists of four separate components (cf. Fig. \ref{fig:gaitLibArchitecture}). The first component, patient-exoskeleton model generation, entails the creation of a patient-specific dynamical model. The patient-exoskeleton model is created by fusing the mass and inertia of each link of a simplified human model with that of each link of the exoskeleton. The simplified human model is created using the patient mass, height, thigh length, and shank length. The human model generation process is based off of the anthropometric data presented in \cite{winter2009biomechanics}. The thigh and shank length of the exoskeleton are adjusted to match those of the patient.

Next, dynamically stable walking gaits are generated for the patient-exoskeleton model using the Partial Hybrid Zero Dynamics (PHZD) method \cite{ames2014human, hereid20163d, Agrawal2017First, gurriet2018towards, harib2018feedback, westervelt2018feedback}. Multiple gaits are generated over a grid of parameters such as patient mass, patient height, step length, step duration, etc. These gaits comprised together form a gait library which is then fitted using a neural network. Once trained, the neural network takes the parameters as inputs, and outputs a joint-level trajectory for each of the 12 joints. The final component of the baseline walking approach is tracking of the joint-level trajectories which is achieved through basic PID control. A deadbeat is implemented to account for early impacts. The desired trajectory each joint is tracking is given by:
\begin{equation}
    q_{des}(t) = q_{nom}(t-t_i) + s(t-t_i)
\end{equation}
where $t_i$ is the time of the latest impact and $s(t)$ is a cubic polynomial satisfying:
\begin{equation}
\left\{
    \begin{array}{l}
     ~ s(0) = q(t_i)-q_{nom}(0), \ s(\alpha T) = 0\\
     s'(0) = q'(t_i)-q_{nom}'(0), \ s'(\alpha T) = 0
    \end{array}
\right.
\end{equation}
with $T$ the nominal duration of a step and $\alpha$ a scalar between 0 and 1.

Additional features were also implemented to improve the performance of nominal exoskeleton walking on hardware. First, flat-foot ankle control was implemented to to ensure that the swing foot always remain parallel to the ground. This ensures a horizontal foot at impact regardless of the time of impact. The flat-foot controller works by using inverse kinematics based on the swing leg tibia IMU to find the ankle joint angles that result in the foot being horizontal. These new swing ankle joint targets are then tracked by the same PID controller as the rest of the joints. Finally, a one degree offset was also added to sagittal ankle desired trajectories to compensate for the effect of hardware flexibilities. 

\subsection{Variable Assistance Architecture}
\label{subsec:assistArchitecture}

\begin{figure*}[t]
    \centering
    \includegraphics[width=\textwidth]{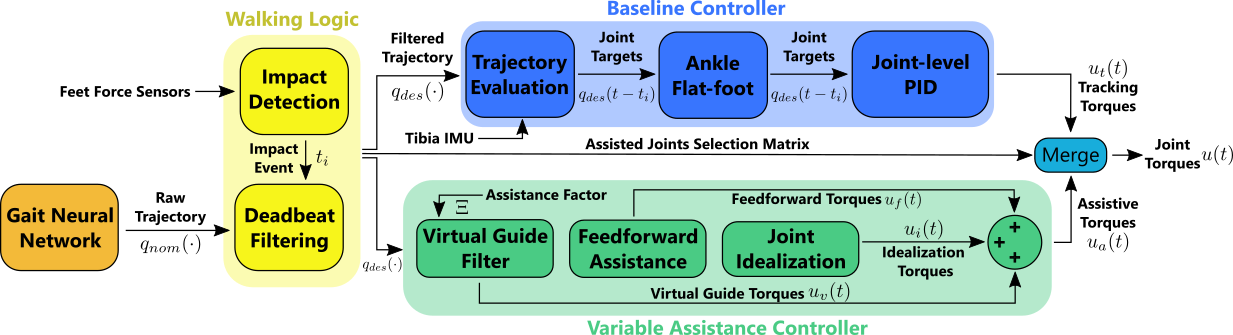}
    \caption{Architecture of the variable assistance framework.}
    \label{fig:assistArchitecture}
\end{figure*}

As discussed in \cite{cai2006implications}, the correct muscle activation pattern is an important criterion for the spinal learning process. It is also showed in the same work that having rigid tracking of the desired gait is sub-optimal in that regard. Leaving some room for the patient to be the one doing the movement yields better results. However, with lower body exoskeletons like Atalante, there is a strong constraint of stability, which limits how much freedom of movement can be given to the user. 

To that end, we explore an approach to precisely control how much freedom is granted to the user, as the better the motoricity of the patient is, the more he or she can be relied on to execute a stable walking pattern. First, we chose joints that we want to let the user control: the assisted joints. All the other joints will be rigidly controlled as described in Sec. \ref{subsec:baselineWalking}. In this work, we choose to only assist the sagittal hip and sagittal knee of the swing leg (cf. Fig. \ref{fig:hardwarePicture}).

The architecture of the variable assistance framework, as shown in Fig. \ref{fig:assistArchitecture}, contains four main components. First, a trajectory is obtained from the neural network based on patient-specific model and desired gait parameters. This trajectory is modulated by the deadbeat mechanism describe in Sec. \ref{subsec:baselineWalking}. This deadbeat mechanism is critical in this case because the nominal joint trajectory will not be followed very accurately when the user is in control of the assisted joints.

The filtered trajectory $q_{des}(\cdot)$ is then fed into two separate controllers. One is the baseline controller presented in Sec. \ref{subsec:baselineWalking}. This controller plays back the trajectory and generates position and velocity targets $q_{des}(t-t_i)$ and $q_{des}'(t-t_i)$ for the PID controllers that in turn generate tracking torques $u_t(t)$. The flatfoot ankle controller separately computes targets for the swing leg ankle that are then substituted in place of the nominal ones.

The other controller is the variable assistance controller. This controller is the heart of the variable assistance framework. The variable assistance controller has three subcomponents: joint idealization, feedforward assistance and virtual guide filter. The torques of these three subcomponents are summed together to form a holistic ``assistive torque'':
\begin{align*}
    u_a(t) = u_i(t)+u_f(t)+u_v(t)
\end{align*}

\newsec{Joint Idealization} The joint idealization component computes the torques required to compensate for gravity and friction in the assisted joints. The goal is to make these joints as transparent as possible such that when there is no assistance, the user does not feel any resistance that would impede his ability to walk freely. The idealization torques are given by:
\begin{equation}
    u_i(t) = k_d\text{sign}\left(q(t)\right) +  k_v q(t) + u_g(t)
\end{equation}
where $u_g(t)$ is computed numerically using inverse dynamics on the model of the empty exoskeleton to compensate the effect of gravity. The friction coefficients $k_d$ and $k_v$ were identified experimentally on the hardware. 

\newsec{Feedforward Assistance} The joint idealization component is not sufficient to make the exoskeleton fully transparent as the inertia of the exoskeleton is not compensated for, which makes the user's legs harder to move. The feedforward assistance component therefore provides feedforward torques $u_f(t)$ -- calculated during the PHZD gait generation process \cite{gurriet2018towards} -- to obtain a first order level of compensation for the inertia of the assisted joints. This does not truly compensate for inertia but at least provides enough assistance for the user to move the exoskeleton legs along the desired trajectory. The intensity of both idealization and feedforward components can be adjusted to produce varying levels of user effort.

\newsec{Virtual Guide Filter} The virtual guide filter computes the joint torques $u_v(t)$ required to limit the discrepancy between the actual and desired trajectory of the assisted joints. This discrepancy limit is described by a tube around the desired trajectory: a virtual guide. The use of virtual guides is most common in the field of human robot interaction \cite{otmane2000active}. The shapes and sizes of the virtual guides can be chosen almost arbitrarily. Given a virtual guide shape, we will talk about ``assistance factor'' to describe the width of the virtual guide tube. Specifically, the assistance factor (denoted by $\Xi$) will be inversely proportional to the width of the guide as defined later by Equation \eqref{eq:qbound}. Using this relationship, $\Xi = 0$ will be the lowest level of exoskeleton assistance, and $\Xi = 1$ will be equivalent to the baseline walking controller without any assistance. The inner workings of this virtual guide filter will be presented in more details the next section.

Finally, the impact detection block also records which leg of the exoskeleton is in stance or swing, and generates an "assisted joints selection matrix" that controls which joints are being assisted at a given instant. Only these joints are assigned the assistive torques. The remaining joints are assigned the baseline tracking torques. The merging of these torques comprise the final joint torques $u(t)$ that are commanded to the exoskeleton.

\subsection{Haptic Feedback}
\label{subsec:hapticFeedback}
Real-time haptic feedback is provided to the user in an effort to increase his ability to follow the desired walking gait. The haptic feedback consists of eight small vibration motors that are located on the front and back of the user's thighs and shanks. Since the user only has control over the gait for $\Xi<1$, as $\Xi=1$ is equivalent to the baseline controller, haptic feedback is only given for assistance factors below 1. The level of vibration is a function of the current joint distance to the virtual guide boundary as well as the assistance factor, and is given by:
\begin{align}
    \text{vibration}(t) = \left|\frac{q(t)-q_{des}(t)}{q_{bound}(t)-q_{des}(t)}\right| \left(1- e^{30(\Xi - 1)}\right)
\end{align}
The amplitude of the vibration increases as the user approaches the virtual guide and only occurs on the vibration motor that is located on the side of the joint that matches direction of the tracking error. For example, if the joint is above the desired target, the user feels a vibration on the front of the limb. Alternatively, if the user is below the desired target angle, the user would feel the vibration on the back of the limb. This way, the user has a physical intuition for where the joints are with respect to the virtual guides.      
\section{Virtual Guide Filter}
\label{sec:theory}

We now present the methodology underlying our approach to providing variable assistance on lower body exoskeletons:  the {\it Virtual Guide Filter}. The theory discussed draws heavily from the field of safety-critical control \cite{aubin2009viability,blanchini2008set} and control barrier functions \cite{ames2016control, ames2019control,gurriet2018online,gurriet2019scalable}, and will be presented here without proofs.

\subsection{Control Barrier Functions}
\label{subsec:subTangCond}
Let's consider continuous-time affine control systems of the form:
\begin{equation}
\dot{x}=f\left(x\right)+g\left(x\right)u.
\label{eq:controlSystemDef}
\end{equation}

\label{ass:SysAssumptions}
\noindent The functions $f$ and $g$ defined on a compact set $X\subset\mathbb{R}^{n}$ are continuously differentiable. The control policies are restricted to be functions $u:\mathbb{R}^{+}\times X\longrightarrow\mathbb{R}^{m}$ Lipschitz continuous in state over $X$ and piecewise continuous in time over $\mathbb{R}^{+}$. We furthermore define $U\subset\mathbb{R}^{m}$ to be the compact and convex set of admissible inputs for this system, i.e. $\forall\, x\in X$ and $\forall\, t\in\mathbb{R}^{+}$, $u\left(t,x\right)\in U$. Finally, we assume that system \eqref{eq:controlSystemDef} has a unique solution over a time interval $\left[0,T\right]$ for any initial condition $x\left(0\right)\in\text{Int}\left(X\right)$ and with $T>0$.

Let's denote the tube we want the system to stay in for a duration $T$ by $\forall\,t\in[0,T],\ \widetilde{S}(t)\subset X$ and require that it is compact. If $\widetilde{S}$ is chosen arbitrarily, it will most certainly contain states that cannot be visited without leading to the system leaving $\widetilde{S}(t)$ before time $T$. Therefore, in order to be able to steer the system to remain in $\widetilde{S}(t),\ \forall\,t\in[0,T]$, it has to be constrained to stay inside a tube $S$ such that $S(t)\subseteq \widetilde{S}(t),\ \forall\,t\in[0,T]$ and that has the property of being a \textit{viable tube}. Such a subset does not contain any unsafe states and it is therefore possible to guarantee the finite time invariance of this new tube through the use of a local characterization of invariance (cf. \cite{aubin2009viability} for more details). The description function $h$ of such a tube $S$ is called a control barrier function.

For that, let's consider smooth practical sets as defined in \cite{blanchini2008set}. To describe such sets, one only needs to consider a continuously differentiable function $h:[0,T]\times \mathbb{R}^{n}\rightarrow\mathbb{R}$ such that:
\begin{equation}
\begin{aligned}S(t) & =\left\{ x\in\mathbb{R}^{n}\mid\ h(t,x)\geq0\right\} \\
\partial S(t) & =\left\{ x\in S\mid\ h(t,x)=0\right\} .
\end{aligned}
\label{eq:setDef}
\end{equation}
This yields the following result from \cite{ames2016control} (see \cite{ames2019control, blanchini2008set, aubin2009viability} for more details). 

\begin{thm}
\label{thm:viabilityTheorem}Given \eqref{eq:controlSystemDef} and control policy $u(t,x) \in U$ subject to the assumptions above, if for almost all $t\in[0,T]$, for all $x\in S(t)$, and for an extended class $\mathscr{K}$ function $\gamma: \mathbb{R} \to \mathbb{R}$:
\begin{equation}
\frac{\partial h}{\partial x }\left(f(x)+g(x)u(t,x)\right) + \frac{\partial h}{\partial t} \geq -\gamma(h(t,x)),\label{eq:sub-tangentiality-condition}
\end{equation}
then $h$ is a control barrier function, and for all $x(0) \in S(0)$ the solution to \eqref{eq:controlSystemDef} remains in $S$, $x(t) \in S$, for all $t\in[0,T]$.
\end{thm}

From this theorem, it naturally follows that the \emph{regulation map} $U_{S}:[0,T]\times S\rightrightarrows U$ characterises the \emph{set of safe inputs}: 
\begin{align}
&U_{S}(t,x) \triangleq  \\
& \quad \left\{ u\in U\mid \frac{\partial h}{\partial x }\left(f(x)+g(x)u\right) + \frac{\partial h}{\partial t} \geq -\gamma(h(t,x))\right\} \nonumber
\end{align}
and captures the constraint that needs to be enforced on the control action for the system to remain in $S$ until $T$, and therefore remain in $\widetilde{S}$ until $T$. Indeed, $u(t) \in U_{S}(t,x(t)), \ \forall\, t\in[0,T]$ guarantees the finite time invariance of $S$.

\subsection{Safe Backward Image}
\label{subsec:implicitRegulationMap}

Finding an explicit representation of viable tubes for even trivial systems is hard and time consuming. To avoid these complexities, the work presented in \cite{gurriet2018online} proposes to use sets that are implicitly defined as a function of the flow of the system under a backup control law, and evaluate $h(t,x)$ online as needed using numerical methods.

Let $\mathcal{U}$ be the set of all continuously differentiable backup control laws taking values in the set of admissible inputs: $u_{b}:\mathbb{R}^{n}\rightarrow U$. Under the assumptions on the control system, we know that for all $u_{b}\in\mathcal{U}$ there exists a solution to \eqref{eq:controlSystemDef} that is unique and defined until $T$. Therefore, one can define $\phi^{u_{b}}:\left[0,T\right]\times X\rightarrow\mathbb{R}^{n}$ to be the \emph{flow} of \eqref{eq:controlSystemDef} under the control law $u_{b}$. Under these assumptions, the map $\phi_{t}^{u_{b}}:X\rightarrow\mathbb{R}^{n}$ defined by $\phi_{t}^{u}\left(x\right)\triangleq\phi^{u}\left(t,x\right)$ is a homeomorphism of $X$ for all $t\in\left[0,T\right]$ \cite{lee2003smooth}. In that context we can define the following set.
\begin{defn}
The \textbf{safe backward image} of $\widetilde{S}(T)$ is defined to be
the set:
\begin{equation}
\varOmega_{T}^{u_{b}}(t)\triangleq\left\{ x\in X\mid \forall\, \tau \in\left[0,T-t\right],\ \phi_{\tau}^{u_{b}}\left(x\right)\in \widetilde{S}(\tau) \right\} .
\end{equation}
\end{defn}

It is then easy to show that if $\varOmega_{T}^{u_{b}}$ is non empty, it is a viability tube subset of $\widetilde{S}$ and that for all $t\in[0,T]$ and for all $x\in \varOmega_{T}^{u_{b}}(t)$, $u_b(x)\in U_{\varOmega_{T}^{u_{b}}}(t,x)$. Furthermore, the set $\varOmega_{T}^{u_{b}}$ enjoys the following property:
\begin{equation}
    \varOmega_{T}^{u_{b}}(t)=\left\{ x\in \widetilde{S} \mid  \underset{\tau\in\left[0,T-t\right]}{\text{min}} h\circ\phi_{\tau}^{u_{b}}\left(x\right)\geq 0 \right\} .
\end{equation}
We thus obtain the main theoretic result of this paper aimed at synthesizing both control barrier functions and controllers that yield safe behavior. 

\begin{prop}
\label{prop:smoothSelection}
Given a nonlinear control system \eqref{eq:controlSystemDef} with a corresponding backup controller $u_{b}:X\rightarrow U$, the function:
\begin{equation}
h_T^{\varOmega}(t,x) \triangleq \underset{\tau\in\left[0,T-t\right]}{\text{min}} h\circ\phi_{\tau}^{u_{b}}\left(x\right),
\end{equation}
is a control barrier function.  Moreover, given a smooth function $\alpha:[0,T]\times X \times \mathbb{R}\rightarrow U$, the control law defined by 
\begin{equation}
     u(t,x)=\alpha\left(t,x,h_T^{\varOmega}(t,x)\right)
\end{equation}
is a smooth selection of $U_{\varOmega_{T}^{u_{b}}}$ if $\alpha\left(t,x,0\right) = u_b(t,x)$, and therefore if $x(0) \in \varOmega_{T}^{u_{b}}(0)$ the system will remain in $\varOmega_{T}^{u_{b}}$, and thus in $\widetilde{S}$ under such a control law.
\end{prop}

To be able to evaluate that policy online, one only has to be able to evaluate the flow of the system $\phi_{\tau}^{u_{b}}$ for all $\tau \in [0,T-t]$. Even though this cannot be done numerically, it can be approximated by numerically integrating the dynamics and evaluating the flow on a finite set of points in $[0,T-t]$ (see \cite{gurriet2018online,gurriet2019scalable}).
Let's now specialize these results for our specific application.
\subsection{Application to Joint Based Filtering}
\label{subsec:safetyFilterApplication}

As presented in Sec. \ref{sec:problemFormulation}, each joint is idealized and handled independently. We therefore consider the following dynamics for each joint:
\begin{equation}
\label{eq:DIdyn}
    J \ddot{q} = u_v + u_f(t-t_i) + u_{ext},
\end{equation}

\noindent where $J$ is the inertia at the joint, $u_v$ is the torque the virtual guide filter can apply, $u_f(t)$ is the feedforward torque applied to the joint, and $u_{ext}$ is the torque applied by the user on the joint. The state of the system is therefore $x=\left[q,\dot{q}\right]\top$.

The virtual guide $\widetilde{S}$ we want to constrain the joint to stay in is characterized by:
\begin{equation}
    h(t,x) = 1-\left(\frac{q_{des}(t-t_i)-q(t)}{q_{bound}(t-t_i)}\right)^2
\end{equation}
for some properly chosen $q_{bound}$ to achieve the desired shape of the guide (cf. Fig. \ref{fig:expVariousShapes} for examples of shapes).

Because $u_{ext}$ is not known ahead of time, a robust version of the results presented before has to be actually used. These extensions are straightforward and will not be presented here due to space constraints, but one must note that they can be used here because system \eqref{eq:DIdyn} is monotone \cite{angeli2003monotone}. In this case, the safe backward image is characterized by:
\begin{equation}
    h_T^{\varOmega}(t,x)=\underset{\substack{\tau\in\left[0,T-t\right] \\ u_{ext}\in\left\{u_{ext}^{min},\ u_{ext}^{max}\right\}}}{\text{min}} h\circ\phi_{\tau}^{u_{b},u_{ext}}\left(q\right),
\end{equation}
where $u_{ext}^{min}$ and $u_{ext}^{max}$ are the extreme values of the disturbance the user can generate. These values were determined experimentally by measuring the maximum joint torque that subjects could generate. So in order to evaluate $h_T^{\varOmega}(t,q)$, the numerical integration of the dynamics only has to be performed twice each time assuming the extremal values of the disturbance. 
The backup policy is chosen to be:
\begin{align}
     u_{b}(t,x)=&K_p(q_{des}(t-t_i)-q)
     + K_d(\dot{q}_{des}(t-t_i)-\dot{q}),
\end{align}
for some properly chosen gains $K_p$ and $K_d$. For this work, these gains were chosen to be the same as the ones used for the PIDs of the baseline controller.

Finally, the filtering law is given by:
\begin{equation}
     u_v(t,x)=\left(\lambda(t,q) + \left(1-\lambda(t,q)\right)\lambda_d(t,q)\right)u_b(t,q,\dot{q}),
\end{equation}
where $\lambda(t,q) = \left(1-h_T^{\varOmega}(t,q)\right)^3$ and $\lambda_d(t,q)=\zeta\frac{dh_T^{\varOmega}(t,q(t))}{dt}$ for some derivative gain $\zeta$ as it is easy to verify that for all $t\in[0,T], \ u_v(t)\in U_{\varOmega_{T}^{u_{b}}}(t,x(t))$. The usage of the derivative term $\lambda_d(t,q)$ helps dampen the behavior of the safety filter. The width of the virtual guide is chosen to be correlated with the assistance factor as defined by:
\begin{align}
    q_{bound}=\pm \left( 0.5+7(1-\Xi) \right) \text{degrees}
    \label{eq:qbound}
\end{align}
and $\lambda(t,q)$ is coupled with the assistance factor by $\lambda(t,q) = \left(1+(\Xi^{10}-1)h_T^{\varOmega}(t,q)\right)^3$ in order to avoid high frequency oscillations when the virtual guide gets small.

\section{Experimental Results}
\label{sec:results}


The variable assistance controller was demonstrated in three separate experiments. First, it was tested on the empty exoskeleton to verify its effectiveness for various tube shapes. Second, the entire framework was tested with eight \textbf{able-bodied} human subjects of masses and heights from 58kg to 91kg and 1.62m to 1.93m. Lastly, the framework was tested over a larger set of assistance factors for a single subject.

\begin{figure}[tb]
    \centering
    \includegraphics[width=\columnwidth]{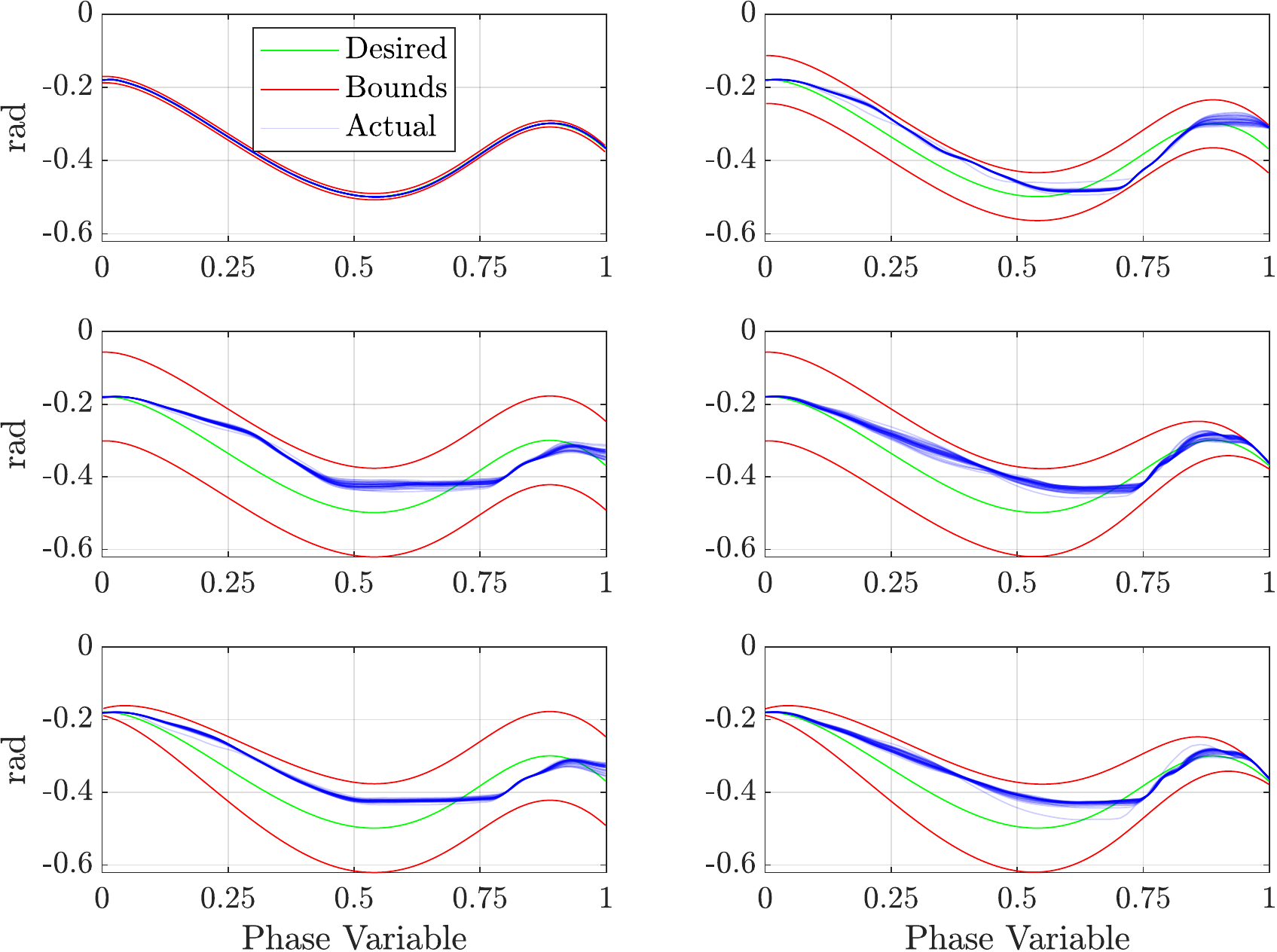}
    \caption{Left hip joint angles for the exoskeleton when empty and hanging in the air. Each plot includes 30 right stance steps and corresponds to a different virtual guide shape.}
    \label{fig:expVariousShapes}
\end{figure}

\newsec{Demonstration of the Virtual Guide Filter} The initial validation experiments were performed on the empty exoskeleton as it hung in the air in an effort to show the behavior of the filter without user perturbations and without feedforward torque. The plots of the experimental results, shown in Fig. \ref{fig:expVariousShapes}, illustrate the actual joint angles over 30 steps with each step overlaid on top of each other. It can be seen that for all tube shapes, the actual joint angles remained inside of the bounds and the filter only acts when necessary.

\begin{figure}[tb]
    \centering
    \includegraphics[width=\columnwidth]{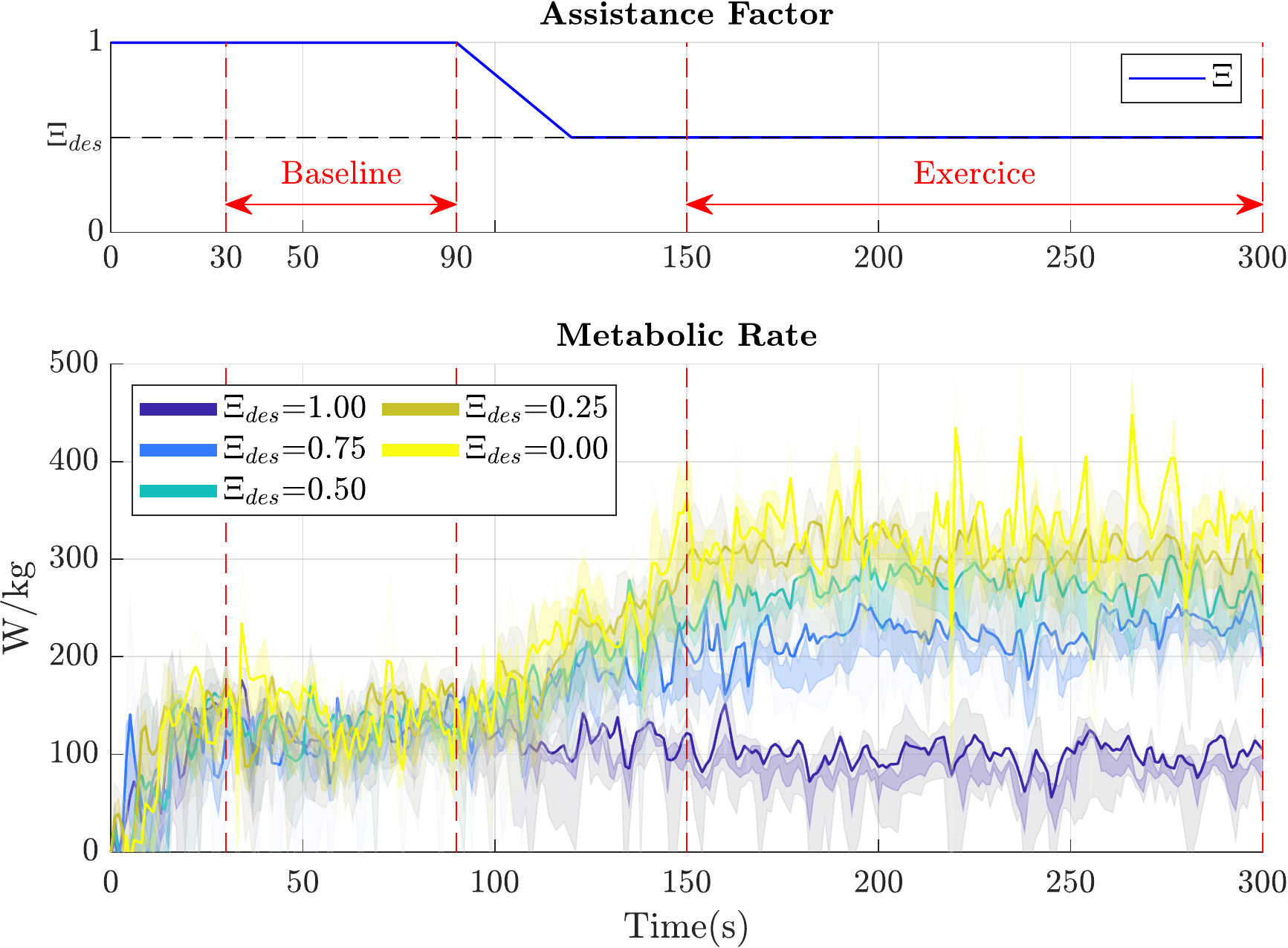}
    \caption{Top figure: variable assistance testing procedure. Bottom figure: metabolic rates as aligned with the testing procedure.}
    \label{fig:expHelpFactorScenario}
\end{figure}
\begin{figure}[tb]
    \centering
    \includegraphics[width=\columnwidth]{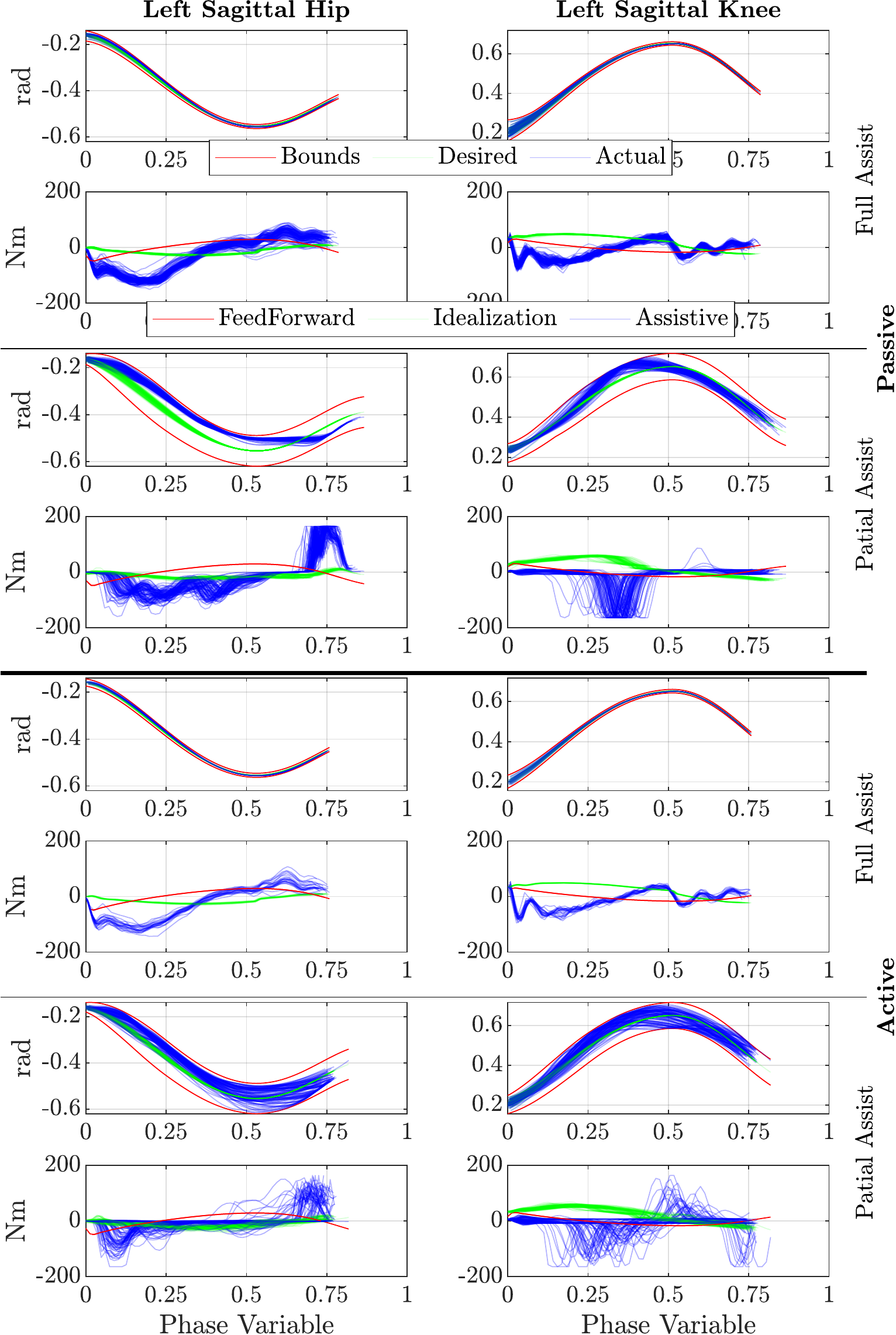}
    \caption{Comparison at the joint level between a subject actively trying to walk and being passive under both full and partial assistance of the exoskeleton. Because of early striking, most steps ended before the phase variable reached 1, unlike in Fig. \ref{fig:expVariousShapes} where the exoskeleton was in the air.}
    \label{fig:expActivePassiveJoints}
\end{figure}

\newsec{Full Assistance versus Partial Assistance} The experimental testing conducted for able-bodied subjects consisted of walking trials lasting five minutes each. The format of each trial is shown in Fig. \ref{fig:expHelpFactorScenario} and is as follows. First, 90 seconds of walking with full assistance, then 30s of transitioning to the desired level of assistance and finally 180s of walking at that desired assistance factor. ``Full Assistance'' corresponds to an assistance factor $\Xi=1$, which is equivalent to the baseline controller without the proposed framework. ``Partial Assistance'' corresponds to $\Xi=0.5$, i.e. $q_{bound}=\pm 4\text{deg}$ (cf. Fig \ref{fig:expActivePassiveJoints}). Beside the subject model parameters, the gait parameters were the same for all subjects. The step length and duration were chosen to be 0.16m and 0.8s respectively.
 
In order to demonstrate the effectiveness of the framework, four trials were conducted per subject. The first two trials were one with Full Assistance and one with Partial Assistance, where the subjects were asked to be completely passive and let the exoskeleton do the work. The same two trials were then repeated but this time asking the subjects to: "Do whatever feels necessary to track the nominal gait".

The required assistive torque, as well as the trajectory tracking, for the four trials of one subject are presented in Fig. \ref{fig:expActivePassiveJoints}. It can be observed that when the subject is passive under partial assistance, the joint trajectories tend to group near the virtual guides as expected. Alternatively, when the subject is active under partial assistance the actual joint trajectories tend to span more of the virtual guide as the subject is actively trying to avoid hitting the bounds of the guide. In all cases, the trajectories stay contained within the virtual guides.

Human metabolic expenditure was recorded for all subjects as it provides critical insight into how much effort the user is exerting. The metabolic rate was determined from oxygen and carbon dioxide exchange rates as measured by a COSMED K4b2. The exchange rates were converted to a metabolic rate using the equation developed by Brockway et al. \cite{brockway1987derivation}. When calculating the metabolic rate, the average metabolic rate recorded over the baseline part of every trial was subtracted from the average rate of the exercise part to isolate the part of the total metabolic power used for compensating for the varying levels of assistance.

\begin{figure}[tb]
    \centering
    \includegraphics[width=\columnwidth]{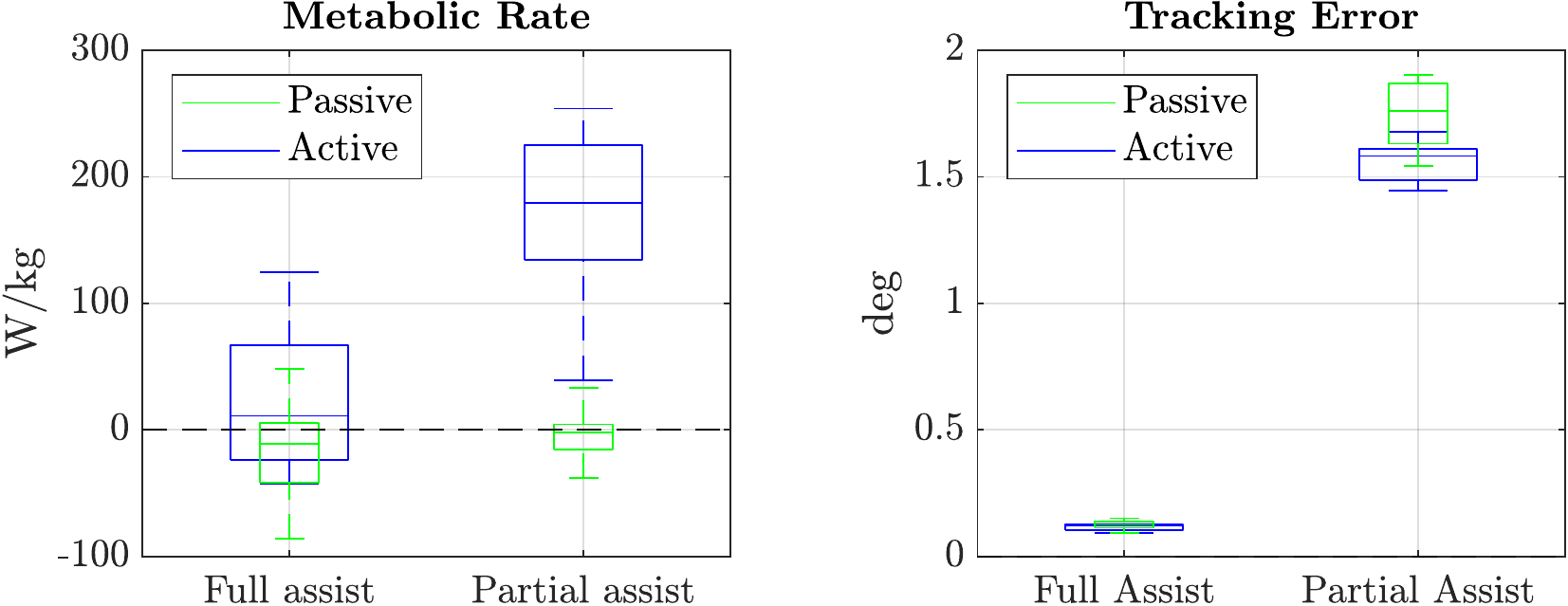}
    \caption{Comparison between tracking accuracy and subject power consumption. The passive data correspond to the subject not doing anything. The active data correspond to the subject trying to follow the nominal gait. Full assist correspond to nominal PID control around the gait, whereas partial assist corresponds to $\pm 4\text{deg}$ wide virtual guides.}
    \label{fig:expActivePassiveCom}
\end{figure}

The results for all eight subjects are summarized in Fig. \ref{fig:expActivePassiveCom}. This figure shows that when the subject was passive, the metabolic rate remained consistent between full assistance and partial assistance. The metabolic rate when passive also is consistently lower than the metabolic rate of the subjects when active at partial assistance. An interesting observation is that the metabolic rate of the subjects when active at full assistance is not much different from that of the subjects when passive. This suggests that the subjects do not feel the need to provide more energy than necessary when the exoskeleton is already providing full assistance. On the other hand, partial assistance incentivises users to contribute to the tracking of the gait which translates into an increase in metabolic rate as expected. Finally, note that on average, the subjects were able to improve the accuracy of tracking in Partial Assist when actively trying.

\newsec{Varying Assistance Factors for One Subject} The testing procedure for the final experiment was the same as discussed previously and shown in Fig. \ref{fig:expHelpFactorScenario} but was repeated with a larger set of assistance factors. The trials were done, in order, with assistance factors $\Xi \in \{1.0, 0.75, 0.5, 0.25, 0.0, 0.25, 0.5, 0.75, 1.0\}$.  A five minute break was taken in between each trial to let the subject return to a resting metabolic rate. The subject also completed one five minute trial while walking on the treadmill at the same velocity as during the trials to compare the subject's nominal walking metabolic rate with that of the exoskeleton testing. This entire procedure was repeated on three consecutive days with the same subject.

\begin{figure}[tb]
    \centering
    \includegraphics[width=\columnwidth]{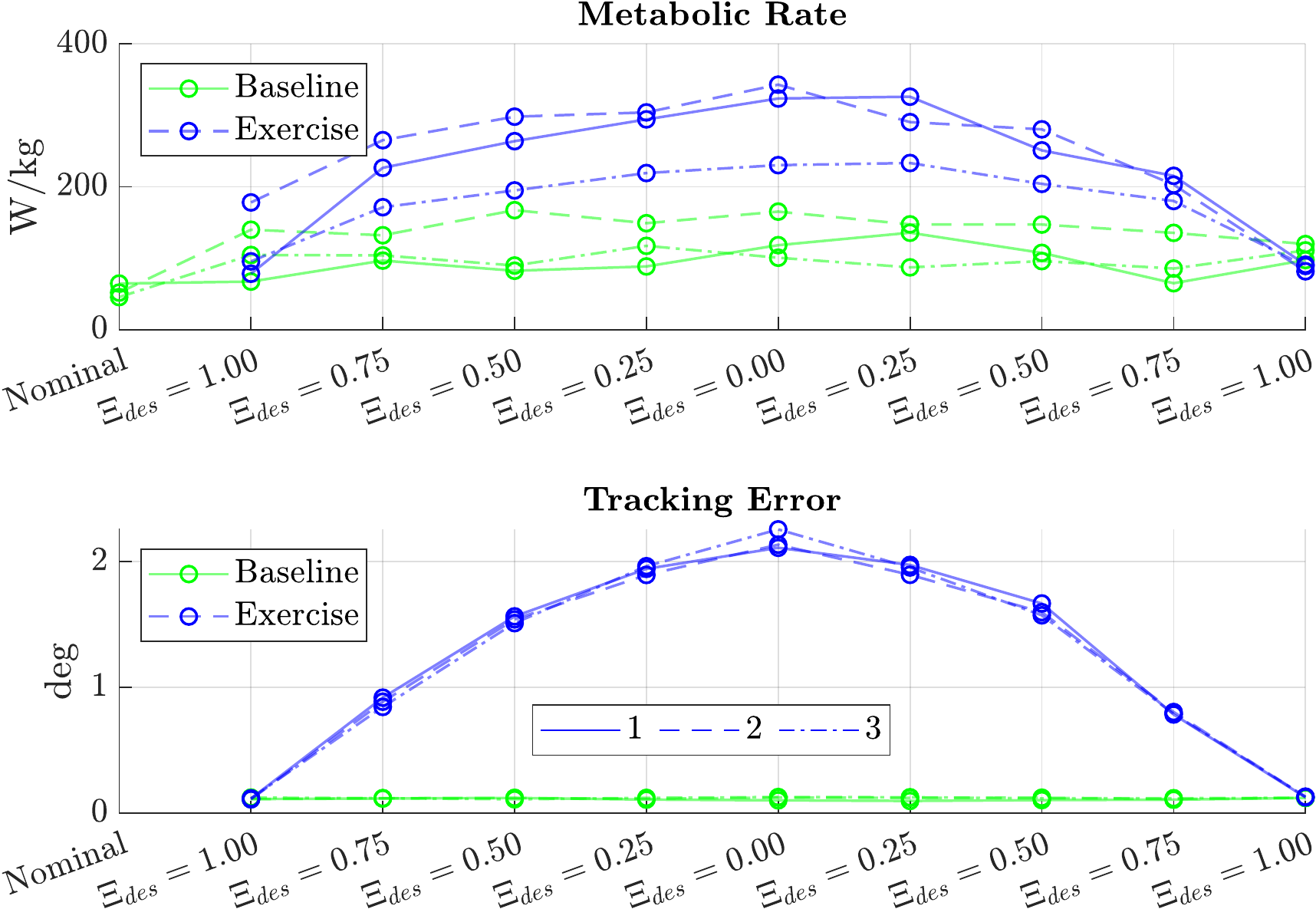}
    \caption{Raw metabolic rate and tracking error in chronological order for the baseline and exercise segments as defined in Fig. \ref{fig:expHelpFactorScenario}. The step length and duration are respectively 0.16m and 0.8s.}
    \label{fig:expHelpFactorUpDown}
\end{figure}

The metabolic power consumption as well as the average tracking error for each segment is reported in Fig. \ref{fig:expHelpFactorUpDown}. The subject's average resting oxygen and carbon dioxide exchange rates, measured at the start of testing, are subtracted from the recorded exchange rates of each trial. Interestingly, it can be seen that the baseline metabolic rate is relatively consistent between all trials and that the data is symmetric around the 0.00 assistance factor trial. This confirms that the increase in exercise metabolic rate for lower assistance factors is due to the lowered assistance and not exhaustion of the subject.

\begin{figure}[tb]
    \centering
    \includegraphics[width=\columnwidth]{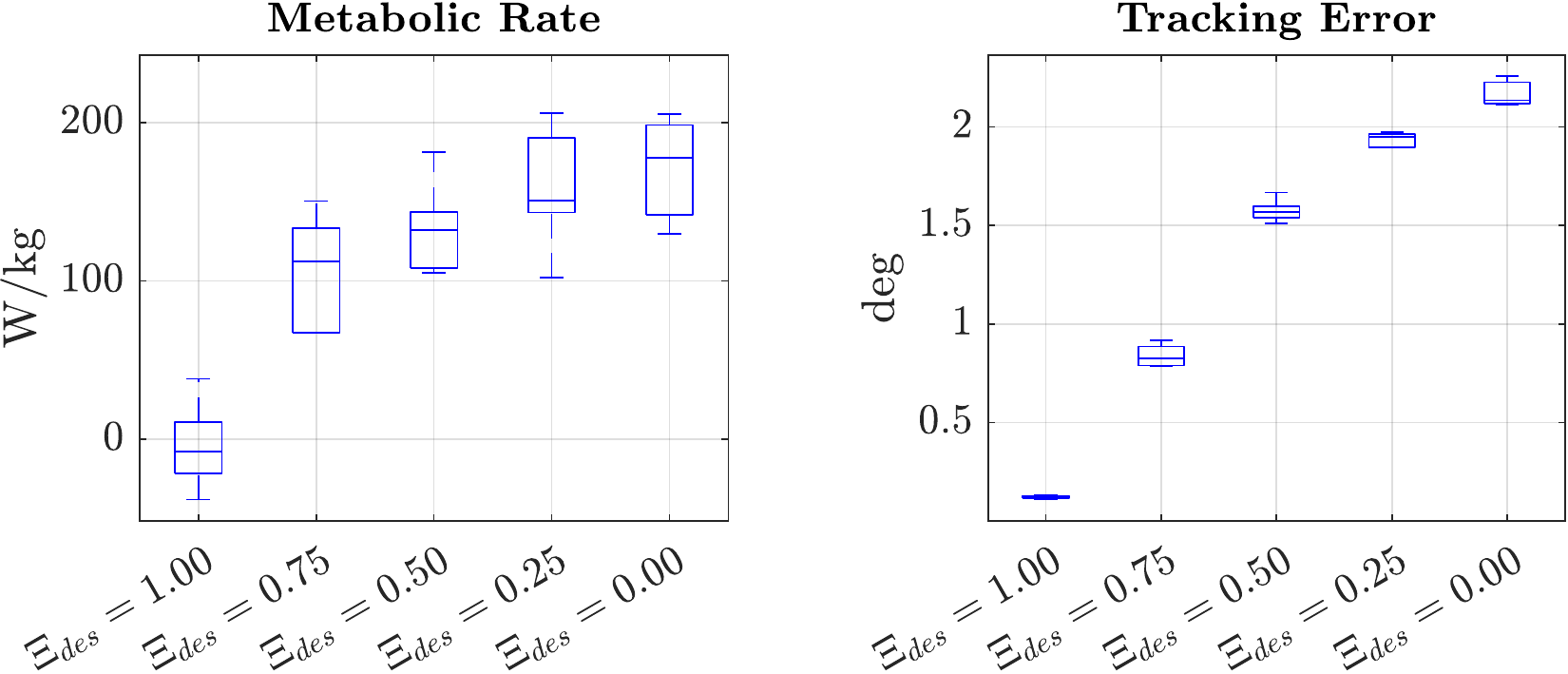}
    \caption{Comparison between tracking accuracy and subject normalized power consumption.}
    \label{fig:expHelpFactor}
\end{figure}

Fig. \ref{fig:expHelpFactor} presents the metabolic rates of the exercise part normalized by the baseline ones for the different values of assistance factor, as well as the corresponding tracking errors. These normalized values indicate a clear trend: \textit{The normalized metabolic rate and the normalized tracking error increase as the assistance factor decreases. }

\section{Conclusion}
\label{sec:conclusion}

In this paper we have presented and demonstrated a framework to achieve variable assistance with a lower body exoskeleton. This result was achieved using tools from controlled set invariance that yield performance guarantees via a virtual guide filter. Through experimental results (cf. video \cite{video}), it was found that the size of the virtual guide had a direct correlation with the amount of power subjects had to provide to track a nominal gait. The authors would also like to note a few additional takeaways. First, it was found that as the assistance factor $\Xi$ was decreased, the operator standing behind the exoskeleton had to help with the stability of the exoskeleton. The authors plan on addressing this issue by adding an active stabilization layer into the framework. Second, it was hard for subjects to track the generated gaits. This is mostly attributed to the fact that the gaits are not very anthropomorphic. Additionally, the proposed haptic feedback was sometimes hard to interpret and act upon by the user, so better feedback strategies are being investigated. Future work includes adapting the walking gaits to match that of the user, as well as investigating how the variable assistance can be used in a true clinical rehabilitation setup.              

\section*{Acknowledgment}

The authors would like to thank the participants that took part in the presented experiments. The authors would also like to thank the entire Wandercraft team that designed Atalante and continues to provide technical support for this project.

\FloatBarrier

\bibliographystyle{IEEEtran}
\bibliography{exo_references}

\begin{thebibliography}{10}
\providecommand{\url}[1]{#1}
\csname url@samestyle\endcsname
\providecommand{\newblock}{\relax}
\providecommand{\bibinfo}[2]{#2}
\providecommand{\BIBentrySTDinterwordspacing}{\spaceskip=0pt\relax}
\providecommand{\BIBentryALTinterwordstretchfactor}{4}
\providecommand{\BIBentryALTinterwordspacing}{\spaceskip=\fontdimen2\font plus
\BIBentryALTinterwordstretchfactor\fontdimen3\font minus
  \fontdimen4\font\relax}
\providecommand{\BIBforeignlanguage}[2]{{%
\expandafter\ifx\csname l@#1\endcsname\relax
\typeout{** WARNING: IEEEtran.bst: No hyphenation pattern has been}%
\typeout{** loaded for the language `#1'. Using the pattern for}%
\typeout{** the default language instead.}%
\else
\language=\csname l@#1\endcsname
\fi
#2}}
\providecommand{\BIBdecl}{\relax}
\BIBdecl

\bibitem{HerrActiveOrthoses}
A.~M. Dollar and H.~Herr, ``Active orthoses for the lower-limbs: challenges and
  state of the art,'' in \emph{2007 IEEE 10th International Conference on
  Rehabilitation Robotics}.\hskip 1em plus 0.5em minus 0.4em\relax IEEE, 2007,
  pp. 968--977.

\bibitem{chen2015design}
B.~Chen, H.~Ma, L.-Y. Qin, X.~Guan, K.-M. Chan, S.-W. Law, L.~Qin, and W.-H.
  Liao, ``Design of a lower extremity exoskeleton for motion assistance in
  paralyzed individuals,'' in \emph{2015 IEEE International Conference on
  Robotics and Biomimetics}.\hskip 1em plus 0.5em minus 0.4em\relax IEEE, 2015,
  pp. 144--149.

\bibitem{esquenazi2012rewalk}
A.~Esquenazi, M.~Talaty, A.~Packel, and M.~Saulino, ``The rewalk powered
  exoskeleton to restore ambulatory function to individuals with thoracic-level
  motor-complete spinal cord injury,'' \emph{American journal of physical
  medicine \& rehabilitation}, vol.~91, no.~11, pp. 911--921, 2012.

\bibitem{neuhaus2011design}
P.~D. Neuhaus, J.~H. Noorden, T.~J. Craig, T.~Torres, J.~Kirschbaum, and J.~E.
  Pratt, ``Design and evaluation of mina: A robotic orthosis for paraplegics,''
  in \emph{2011 IEEE International Conference on Rehabilitation
  Robotics}.\hskip 1em plus 0.5em minus 0.4em\relax IEEE, 2011, pp. 1--8.

\bibitem{swift2011control}
T.~A. Swift, ``Control and trajectory generation of a wearable mobility
  exoskeleton for spinal cord injury patients,'' Ph.D. dissertation, UC
  Berkeley, 2011.

\bibitem{krut2010moonwalker}
S.~Krut, M.~Benoit, E.~Dombre, and F.~Pierrot, ``Moonwalker, a lower limb
  exoskeleton able to sustain bodyweight using a passive force balancer,'' in
  \emph{2010 IEEE International Conference on Robotics and Automation}.\hskip
  1em plus 0.5em minus 0.4em\relax IEEE, 2010, pp. 2215--2220.

\bibitem{zoss2006biomechanical}
A.~B. Zoss, H.~Kazerooni, and A.~Chu, ``Biomechanical design of the berkeley
  lower extremity exoskeleton (bleex),'' \emph{IEEE/ASME Transactions on
  mechatronics}, vol.~11, no.~2, pp. 128--138, 2006.

\bibitem{jimenez2012review}
R.~Jimenez-Fabian and O.~Verlinden, ``Review of control algorithms for robotic
  ankle systems in lower-limb orthoses, prostheses, and exoskeletons,''
  \emph{Medical engineering \& physics}, vol.~34, no.~4, pp. 397--408, 2012.

\bibitem{tucker2015control}
M.~R. Tucker, J.~Olivier, A.~Pagel, H.~Bleuler, M.~Bouri, O.~Lambercy, J.~del
  R~Mill{\'a}n, R.~Riener, H.~Vallery, and R.~Gassert, ``Control strategies for
  active lower extremity prosthetics and orthotics: a review,'' \emph{Journal
  of neuroengineering and rehabilitation}, vol.~12, no.~1, p.~1, 2015.

\bibitem{anam2012active}
K.~Anam and A.~A. Al-Jumaily, ``Active exoskeleton control systems: State of
  the art,'' \emph{Procedia Engineering}, vol.~41, pp. 988--994, 2012.

\bibitem{gurriet2018towards}
T.~Gurriet, S.~Finet, G.~Boeris, A.~Duburcq, A.~Hereid, O.~Harib, M.~Masselin,
  J.~Grizzle, and A.~D. Ames, ``Towards restoring locomotion for paraplegics:
  Realizing dynamically stable walking on exoskeletons,'' in \emph{2018 IEEE
  International Conference on Robotics and Automation (ICRA)}.\hskip 1em plus
  0.5em minus 0.4em\relax IEEE, 2018, pp. 2804--2811.

\bibitem{cai2006implications}
L.~L. Cai, A.~J. Fong, C.~K. Otoshi, Y.~Liang, J.~W. Burdick, R.~R. Roy, and
  V.~R. Edgerton, ``Implications of assist-as-needed robotic step training
  after a complete spinal cord injury on intrinsic strategies of motor
  learning,'' \emph{Journal of Neuroscience}, vol.~26, no.~41, pp.
  10\,564--10\,568, 2006.

\bibitem{srivastava2014assist}
S.~Srivastava, P.-C. Kao, S.~H. Kim, P.~Stegall, D.~Zanotto, J.~S. Higginson,
  S.~K. Agrawal, and J.~P. Scholz, ``Assist-as-needed robot-aided gait training
  improves walking function in individuals following stroke,'' \emph{IEEE
  Transactions on Neural Systems and Rehabilitation Engineering}, vol.~23,
  no.~6, pp. 956--963, 2014.

\bibitem{cai2006assist}
L.~L. Cai, A.~J. Fong, Y.~Liang, J.~Burdick, and V.~R. Edgerton,
  ``Assist-as-needed training paradigms for robotic rehabilitation of spinal
  cord injuries,'' in \emph{2006 IEEE International Conference on Robotics and
  Automation (ICRA)}.\hskip 1em plus 0.5em minus 0.4em\relax IEEE, 2006, pp.
  3504--3511.

\bibitem{zanotto2014adaptive}
D.~Zanotto, P.~Stegall, and S.~K. Agrawal, ``Adaptive assist-as-needed
  controller to improve gait symmetry in robot-assisted gait training,'' in
  \emph{2014 IEEE international conference on robotics and automation
  (ICRA)}.\hskip 1em plus 0.5em minus 0.4em\relax IEEE, 2014, pp. 724--729.

\bibitem{aubin2009viability}
J.-P. Aubin, \emph{Viability theory}.\hskip 1em plus 0.5em minus 0.4em\relax
  Springer Science, 2009.

\bibitem{blanchini2008set}
F.~Blanchini and S.~Miani, \emph{Set-theoretic methods in control}.\hskip 1em
  plus 0.5em minus 0.4em\relax Springer, 2008.

\bibitem{ames2016control}
A.~D. Ames, X.~Xu, J.~W. Grizzle, and P.~Tabuada, ``Control barrier function
  based quadratic programs for safety critical systems,'' \emph{IEEE
  Transactions on Automatic Control}, vol.~62, no.~8, pp. 3861--3876, 2017.

\bibitem{ames2019control}
A.~D. Ames, S.~Coogan, M.~Egerstedt, G.~Notomista, K.~Sreenath, and P.~Tabuada,
  ``Control barrier functions: Theory and applications,'' \emph{arXiv preprint
  arXiv:1903.11199}, 2019.

\bibitem{agrawal2017discrete}
A.~Agrawal and K.~Sreenath, ``Discrete control barrier functions for
  safety-critical control of discrete systems with application to bipedal robot
  navigation.'' in \emph{Robotics: Science and Systems}, 2017.

\bibitem{wang2017safety}
L.~Wang, A.~D. Ames, and M.~Egerstedt, ``Safety barrier certificates for
  collisions-free multirobot systems,'' \emph{IEEE Transactions on Robotics},
  vol.~33, no.~3, pp. 661--674, 2017.

\bibitem{winter2009biomechanics}
D.~A. Winter, \emph{Biomechanics and motor control of human movement}.\hskip
  1em plus 0.5em minus 0.4em\relax John Wiley \& Sons, 2009.

\bibitem{ames2014human}
A.~D. Ames, ``Human-inspired control of bipedal walking robots,'' \emph{IEEE
  Transactions on Automatic Control}, vol.~59, no.~5, pp. 1115--1130, 2014.

\bibitem{hereid20163d}
A.~Hereid, E.~A. Cousineau, C.~M. Hubicki, and A.~D. Ames, ``3{D} dynamic
  walking with underactuated humanoid robots: A direct collocation framework
  for optimizing hybrid zero dynamics,'' in \emph{2016 IEEE International
  Conference on Robotics and Automation (ICRA)}.\hskip 1em plus 0.5em minus
  0.4em\relax IEEE, 2016, pp. 1447--1454.

\bibitem{Agrawal2017First}
A.~Agrawal, O.~Harib, A.~Hereid, S.~Finet, M.~Masselin, L.~Praly, A.~D. Ames,
  K.~Sreenath, and J.~W. Grizzle, ``First steps towards translating {HZD}
  control of bipedal robots to decentralized control of exoskeletons,''
  \emph{{IEEE} Access}, vol.~5, pp. 9919--9934, 2017.

\bibitem{harib2018feedback}
O.~Harib, A.~Hereid, A.~Agrawal, T.~Gurriet, S.~Finet, G.~Boeris, A.~Duburcq,
  M.~E. Mungai, M.~Masselin, A.~D. Ames \emph{et~al.}, ``Feedback control of an
  exoskeleton for paraplegics: Toward robustly stable, hands-free dynamic
  walking,'' \emph{IEEE Control Systems Magazine}, vol.~38, no.~6, pp. 61--87,
  2018.

\bibitem{westervelt2018feedback}
E.~R. Westervelt, J.~W. Grizzle, C.~Chevallereau, J.~H. Choi, and B.~Morris,
  \emph{Feedback control of dynamic bipedal robot locomotion}.\hskip 1em plus
  0.5em minus 0.4em\relax CRC press, 2018.

\bibitem{otmane2000active}
S.~Otmane, M.~Mallem, A.~Kheddar, and F.~Chavand, ``Active virtual guides as an
  apparatus for augmented reality based telemanipulation system on the
  internet,'' in \emph{Proceedings 33rd Annual Simulation Symposium (SS
  2000)}.\hskip 1em plus 0.5em minus 0.4em\relax IEEE, 2000, pp. 185--191.

\bibitem{gurriet2018online}
T.~Gurriet, M.~Mote, A.~D. Ames, and E.~Feron, ``An online approach to active
  set invariance,'' in \emph{2018 IEEE Conference on Decision and Control
  (CDC)}.\hskip 1em plus 0.5em minus 0.4em\relax IEEE, 2018, pp. 3592--3599.

\bibitem{gurriet2019scalable}
T.~Gurriet, M.~Mote, A.~Singletary, E.~Feron, and A.~D. Ames, ``A scalable
  controlled set invariance framework with practical safety guarantees,'' in
  \emph{2019 IEEE Conference on Decision and Control (CDC)}.\hskip 1em plus
  0.5em minus 0.4em\relax IEEE, 2019, p. n/a.

\bibitem{lee2003smooth}
J.~M. Lee, ``Smooth manifolds,'' in \emph{Introduction to Smooth
  Manifolds}.\hskip 1em plus 0.5em minus 0.4em\relax Springer, 2003, pp. 1--29.

\bibitem{angeli2003monotone}
D.~Angeli and E.~D. Sontag, ``Monotone control systems,'' \emph{IEEE
  Transactions on Automatic Control}, vol.~48, no.~10, pp. 1684--1698, 2003.

\bibitem{brockway1987derivation}
J.~Brockway, ``Derivation of formulae used to calculate energy expenditure in
  man.'' \emph{Human nutrition. Clinical nutrition}, vol.~41, no.~6, pp.
  463--471, 1987.

\bibitem{video}
``Video of the experimental results.'' \url{https://youtu.be/UJC5j4BFxyo}.

\end{thebibliography}

\end{document}